\pdfoutput=1

\documentclass[11pt]{article}

\usepackage[final]{acl}

\usepackage{times}
\usepackage{latexsym}

\usepackage[T1]{fontenc}

\usepackage[utf8]{inputenc}

\usepackage{microtype}

\usepackage{inconsolata}

\usepackage{graphicx}
\usepackage{booktabs}
\usepackage{multirow}
\usepackage{algorithm}
\usepackage{algorithmic}
\usepackage{amsmath} %
\usepackage{smile}
\usepackage{cleveref}
\usepackage{amssymb}  %
\usepackage{booktabs}
\usepackage{pifont}
\newcommand{\xmark}{\ding{55}} %

\usepackage{makecell} %
\usepackage{xspace}

\newcommand{\cost}[1]{\colorbox[RGB]{243, 243, 243}{\makebox(30,6){#1}}}

\newcommand{\ours}{\texttt{CNTP}\xspace}

\title{Cautious Next Token Prediction}

\author{Yizhou Wang$^\dag\thanks{Work was done while Yizhou Wang was an intern at Adobe.}$, Lingzhi Zhang$^{\ddag}$, Yue Bai$^\dag$, Mang Tik Chiu$^\ddag$, Zhengmian Hu$^\ddag$,\\ {\bf Mingyuan Zhang$^\dag$, Qihua Dong$^\dag$, Yu Yin$\S$, Sohrab Amirghodsi$^{\ddag}$ and Yun Fu$^\dag$} \\
$^\dag$Northeastern University, $^\ddag$Adobe, $\S$Case Western Reserve University \\
wyzjack990122@gmail.com, yunfu@ece.neu.edu
}

\begin{document}
\maketitle

\begin{abstract}
Next token prediction paradigm has been prevailing for autoregressive models in the era of LLMs. The current default sampling choice for popular LLMs is temperature scaling together with nucleus sampling~\cite{holtzman2019curious} to balance diversity and coherence. Nevertheless, such approach leads to inferior performance in various NLP tasks when the model is not certain about testing questions. To this end, we propose a brand new training-free decoding strategy, dubbed as Cautious Next Token Prediction (\ours). In the decoding process, if the model has comparatively high prediction entropy at a certain step, we sample multiple trials starting from the step independently and stop when encountering any punctuation. Then we select the trial with the lowest perplexity score viewed as the most probable and reliable trial path given the model's capacity. The trial number is negatively correlated with the prediction confidence, i.e., the less confident the model is, the more trials it should sample. This is consistent with human beings' behaviour: when feeling uncertain or unconfident, one tends to think more creatively, exploring multiple thinking paths, to cautiously select the path one feels most confident about. Extensive experiments on both LLMs and MLLMs show that our proposed \ours approach outperforms existing standard decoding strategies consistently by a clear margin. Moreover, the integration of \ours with self consistency~\cite{wang2022self} can further improve over vanilla self consistency. We believe our proposed \ours has the potential to become one of the default choices for LLM decoding. Code is available at \url{https://github.com/wyzjack/CNTP}.
\end{abstract}

\section{Introduction}
\begin{table}[ht]
\centering
\caption{Comparison of \ours with Stochastic Decoding ({\bf SD}), Greedy Decoding ({\bf GD}), and Beam Search ({\bf BS}) on key text generation properties. A \checkmark indicates exhibiting the property, while a \xmark indicates not.}
\label{tab:method_comparison}
\scalebox{0.9}{
\begin{tabular}{lcccc}
\toprule
\textbf{Textual Property} & \textbf{SD} & \textbf{GD} & \textbf{BS} & \textbf{\ours} \\
\midrule
Stochasticity            & \checkmark & \xmark     & \xmark     & \checkmark \\
Coherence                & \xmark     & \checkmark & \checkmark & \checkmark \\
Creativity               & \checkmark & \xmark     & \xmark     & \checkmark \\
Computational efficiency & \checkmark & \checkmark & \xmark     & \checkmark \\
\bottomrule
\end{tabular}
}
\end{table}

Large Language Models (LLMs) have advanced the capabilities of natural language processing rapidly, achieving remarkable performance in tasks spanning machine translation, summarization, and question answering~\citep{gpt3_20,radford2021learning,gpt4,llama_23,llama2_23,llama3,shen2025efficient,zhang2025boosting}. 
Beyond text-only domains, multimodal LLMs (MLLMs) extend these breakthroughs to image and video understanding, yielding transformative results in visual question answering and video event comprehension~\citep{llava,llava1.5,Li_2024_CVPR}. 
Despite this progress, test-time decoding strategies remain a bottleneck: standard approaches (e.g., greedy decoding, top-$k$ sampling, nucleus sampling) can either produce dull or suboptimal completions in uncertain contexts, undermining overall performance~\citep{wang2023selfconsistency,aggarwal-etal-2023-lets}. 
Hence, devising novel inference-time algorithms to more effectively handle ambiguity and maintain coherent reasoning has become a critical challenge.

Recent research aims to bolster the reliability and depth of LLM outputs through chain-of-thought (CoT) reasoning, self-consistency voting, and iterative self-refinement~\citep{wei2022chain,gou2023critic,wang2022self,aggarwal-etal-2023-lets,chen2023universal}. 
These approaches have demonstrated impressive improvements on QA and reasoning benchmarks by explicitly sampling multiple solution paths. 
However, extensive multi-sample strategies often inflate computational cost, while purely single-sample methods risk hallucinations and local optimum traps. 
Furthermore, a great number of self-correction methods rely on external feedback signals~\cite{gou2023critic,chen2023universal}, which can be intractable to deploy at scale. Therefore, a method that dynamically adapts its exploration only when the model is unconfident, and does so efficiently, remains highly desirable for real-world deployment and application. In this paper, we propose a \textbf{Cautious Next Token Prediction (\ours)} approach that selectively samples multiple candidate paths whenever the model’s prediction entropy is high, then automatically chooses the path with the lowest perplexity. 
Through conditioning the sampling depth on confidence, \ours focuses computational resources precisely where the model is most uncertain, leading to stronger results in both purely linguistic and multimodal tasks.  Extensive experiments show that \ours consistently outperforms common decoding approaches. In summary, we make the following contributions:
\begin{itemize}
    \item We introduce \ours, a novel inference-only decoding strategy for LLMs that adaptively samples multiple continuations based on model confidence, thereby achieving high precision without the loss of diversity.
    \item We propose an entropy-based mechanism to control the number of sampled trials, enabling \ours to conserve computational budget while reducing errors systematically  in high-uncertainty regions.
    \item Through comprehensive studies on both language-only and multimodal benchmarks, we demonstrate that \ours obtains superior performance in comparison to widely adopted decoding baselines consistently. In addition, we show how \ours can be combined with self-consistency~\cite{wang2022self}, further enhancing performance steadily.
\end{itemize}

\section{Related works}

\paragraph{LLM Reasoning}
The reasoning ability of Large Language models also plays a crucial role in multi-modality tasks including image~\citep{llava,llava1.5,zhu2024minigpt} and video understanding~\citep{maaz-etal-2024-video,wang2023vaquita,Song_2024_CVPR,Li_2024_CVPR,chen2024through}. LLMs have demonstrated emergent reasoning abilities through chain-of-thought (CoT) prompting, enabling them to surpass earlier approaches on multi-step inference tasks~\citep{wei2022chain,kojima2022large}. 
Self-consistency~\cite{wang2023selfconsistency} extends CoT by sampling multiple reasoning paths and selecting the most frequent solution, yielding significant gains on math and commonsense QA tasks.Adaptive-Consistency~\citep{aggarwal-etal-2023-lets} halts sampling early if partial agreement is reached. The Tree-of-Thought framework explores multiple candidate “thought” sequences in a search-like manner, substantially improving performance on puzzles and longer-horizon tasks~\citep{yao2024tree}. In multimodal contexts, recent works incorporate vision inputs into multi-stage reasoning, achieving state-of-the-art results on image- and video-based QA~\citep{xu2024llava,fei2024video}. Compared with these methods, our approach offers a simpler yet more robust mechanism for stepwise reasoning without relying on extensive prompt engineering or heavy sampling, yielding more effective inference.
\paragraph{LLM Test-time Sampling}
At inference time, widely used stochastic strategies such as top-p sampling or top-$p$ (nucleus) sampling~\cite{holtzman2019curious} balance diversity and fluency, becoming the default choices for modern LLMs particularly in industrial products~\cite{gpt4,guo2025deepseek,team2025kimi,yang2024qwen2}. More recently, min-p sampling~\cite{nguyen2024turning} puts forward a dynamic truncation method which adjusts the sampling threshold contigent on the model’s confidence by scaling based on the top token probability. As to deterministic sampling strategies, Greedy Decoding selects the single most probable token at each step by choosing the one with the maximum token probability. Beam Search~\cite{graves2012sequence} keeps track of a fixed number of top candidate sequences at each step and finally selects the trial with the highest joint probability. Unlike these strategies, our method adaptively samples multiple paths at high-entropy steps, using perplexity to select the best continuation, thereby enhancing answer quality without losing stochasticity.

\paragraph{LLM Self-correction}
LLMs sometimes produce errors or hallucinations, prompting the development of self-refinement or self-correction.~\citet{gou2023critic} shows that LLMs can self-correct by taking advantage of external tools to validate their initial outputs, gathering feedback on specific aspects, and then refining the content based on that feedback iteratively.~\citet{kumar2024training} proposes a finetuning strategy to make LLM self-correct by engaging in multi-turn online reinforcement learning on its own self-generated correction traces. However,~\citet{huang2023large} comes to the conclusion that current LLMs still cannot do self-reflection via directly prompting to itself to reflect on the generated answers without the access to external tools. Our proposed approach remains fully self-contained without extra critic models or extensive domain-specific tuning. In fact, our approach can be viewed as local and progressive self-correction by means of pursuing low sentence perplexity in attempted short explorations.

\section{Method}

\begin{figure*}[tb]
    \centering
    \includegraphics[width=\linewidth]{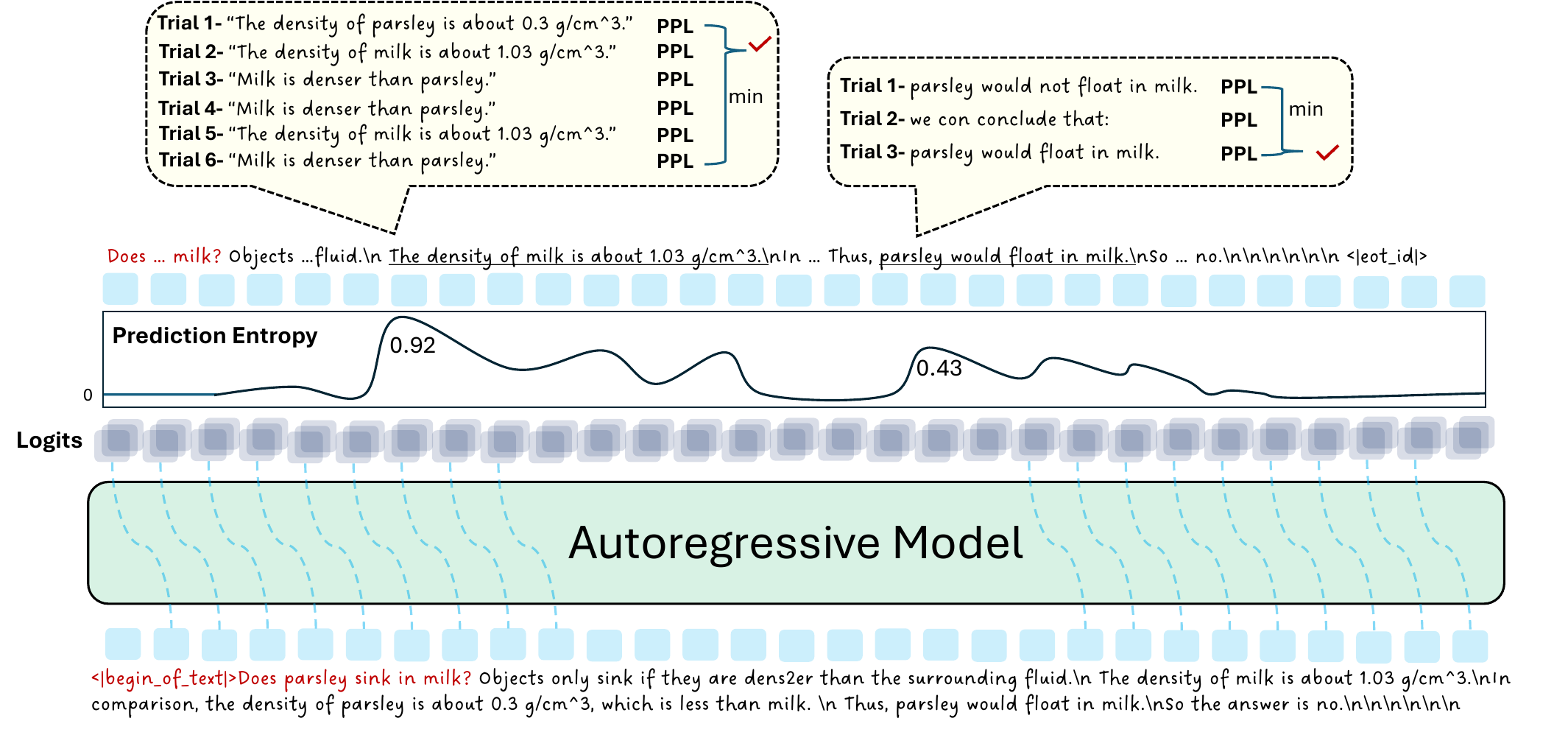}
    \caption{Overview of {\bf C}autious {\bf N}ext {\bf T}oken {\bf P}rediction ({\bf \ours}) method. Given a question in the testing phase, the AR model predict next tokens in an iterative way. Different from traditional one-token-by-one-token generation style, we leverage the prediction entropy of the token probability vectors and sample multiple trials to look ahead if the entropy is high enough. The different trials are sampled independently and stop when encountering a punctuation, then the trial with the lowest perplexity is selected to be deemed the most confident choice. The higher the prediction entropy is, the more trial paths we sample. The QA example is generated by our approach on one of the StrategyQA~\cite{geva2021did} testing samples. The question texts are in {\color{red}{red}} and the answer texts are in black.  }
    \label{fig: cntp_overview}
\end{figure*}

\begin{algorithm}[t]
\caption{Cautious Next Token Prediction}
\label{alg:cntp}
\begin{algorithmic}[1]
    \STATE \textbf{Initialization:} Language model $M$, initial sequence $s$, max trials $N_{\text{max}}$, entropy thresholds $H_{\min}, H_{\max}$
    \WHILE{s does not satisfy stopping criteria}
        \STATE Compute token distribution $p(\cdot \mid s)$ via $M$
        \STATE $H \leftarrow -\sum_{w} p(w \mid s)\,\log p(w \mid s)$ 
        \STATE $N \leftarrow \max\Bigl(1,\,\min\Bigl(N_{\text{max}},\bigl\lfloor \tfrac{H - H_{\min}}{H_{\max} - H_{\min}} \times N_{\text{max}} \bigr\rfloor\Bigr)\Bigr)$
        \IF{$N = 1$}
            \STATE Sample a single token $s_{\text{single}}$
            \STATE $s \!\leftarrow\! s \,+\, s_{\text{single}}$
        \ELSE
            \FOR{$i \!\leftarrow\! 1$ to $N$}
                \STATE Sample path $s_i$ via $M$ until punctuation or satisfying stopping criteria
                \STATE $\mathcal{L}(s_i) \!\leftarrow\! -\sum_{t} \log p\bigl(w_t \!\mid\! s_{<t}; M\bigr)$
                \STATE $\mathrm{PPL}(s_i) \!\leftarrow\!\exp\bigl(\mathcal{L}(s_i)/\lvert s_i\rvert\bigr)$
            \ENDFOR
            \STATE $s \!\leftarrow\! s \,+\, \arg\min_{s_i}\mathrm{PPL}(s_i)$ 
        \ENDIF
    \ENDWHILE
    \RETURN completed sequence $s$
\end{algorithmic}
\end{algorithm}

\subsection{Motivation}
When human beings solve complex tasks---such as proving math theorems---they typically proceed step-by-step, carefully reviewing each line for correctness. If a certain step appears ambiguous or risky, they explore multiple potential pathways, reflect on each, and ultimately choose the route that seems most convincing or “probable.” This meta-cognitive process of careful thinking motivates our proposed Cautious Next Token Prediction (\ours) algorithm. Specifically, in human exam settings, {\bf individuals often re-check key steps and sample alternative reasoning paths when unsure, only finalizing the path that best resonates with previously established facts. Such progressive practice usually results in better human performance}. By analogy, we hypothesize that LLMs might also benefit from conditionally branching into multiple future continuations whenever they sense high uncertainty. Our core insight is to compute an \emph{entropy} measure that indicates how “unsure” the model is, and trigger more thorough exploration exactly at those points. Once possible continuations are sampled, the model’s own likelihood function can judge the best candidate to proceed with, mirroring how humans choose the strongest proof line. In this work, {\bf we are thrilled to confirm the empirical effects of such practice on LLMs as well}.

\subsection{Cautious Next Token Prediction}
\label{sec:cntp}

\paragraph{Preliminaries and Notation.}
Let $M$ be a language model that defines a probability distribution $p_{\theta}(w \mid s)$ over the next token $w$ given the current (partial) sequence $s$. At each generation step, $M$ computes the distribution over its vocabulary $V$ as

\begin{equation}
    \small
    p_{\theta}(w \mid s) = \frac{\exp\bigl(\text{logit}_\theta(s,w)\bigr)}{\sum_{v \in V} \exp\bigl(\text{logit}_\theta(s,v)\bigr)},
\end{equation}
where $\text{logit}_\theta(s,w)$ is the unnormalized score for token $w$. 
We measure uncertainty via the \emph{entropy} of this distribution:
\begin{equation}
    \small
    H(s) = - \sum_{w \in V} p_{\theta}(w \mid s)\,\log p_{\theta}(w \mid s).
\end{equation}
When $H(s)$ is large, the model is more uncertain and less confident about the next token.

\paragraph{Adaptive Trial Sampling.}
As shown in Algorithm~\ref{alg:cntp}, our \ours method adaptively decides how many candidate continuations to explore based on the entropy $H(s)$. We specify two thresholds, $H_{\min}$ and $H_{\max}$, and a maximum trial budget $N_{\text{max}}$. We then map the current entropy $H(s)$ to a suitable number of trials $N$:
\begin{equation}
    \small
    N = \max\Bigl(1,\min\Bigl(N_{\text{max}},\bigl\lfloor \tfrac{(H(s) - H_{\min})\cdot N_{\text{max}} }{(H_{\max}-H_{\min})} \bigr\rfloor\Bigr)\Bigr).
\end{equation}
This ensures $N=1$ when $H(s)$ is below $H_{\min}$ (i.e., the model is quite confident), and $N=N_{\text{max}}$ when $H(s)$ exceeds $H_{\max}$. For each trial $i\in\{1,\dots,N\}$, we \emph{sample a candidate path} $s_i$ until a punctuation token or stopping criterion. We compute the path’s negative log-likelihood (NLL),
\begin{equation}
    \small
    \mathcal{L}(s_i) = -\sum_{t=1}^{|s_i|} \log p_{\theta}\bigl(w_t \mid s_{<t}\bigr),
\end{equation}
and convert it to perplexity (PPL),
\begin{equation}
    \small
    \mathrm{PPL}(s_i)=\exp\bigl(\tfrac{\mathcal{L}(s_i)}{|s_i|}\bigr).
\end{equation}
The path with the \emph{lowest} perplexity is considered the most likely to be correct. \ours thus updates the sequence $s$ by appending the best candidate continuation. The process repeats until a global termination condition (e.g., an end-of-sentence token or length limit) is reached.

\paragraph{Algorithmic Overview.}
Algorithm~\ref{alg:cntp} and Fig.~\ref{fig: cntp_overview} summarizes the entire \ours loop. Notably, the method is \emph{cautious} in that it only expends extra sampling steps when the model signals uncertainty. At confident steps, \ours automatically defaults to a single-sample approach. {\bf This design aligns with our human-inspired motivation: we “think harder” only when the situation is ambiguous.}

\subsection{Complexity Analysis}
Let $L$ be the final sequence length. In the worst-case scenario, \ours samples $N_{\text{max}}$ continuations for each token, producing $O(L \times N_{\text{max}})$ total decoding operations. This upper bound resembles standard multi-sample decoding. However, because $N$ is only large when $H(s)$ is high, \ours often runs with $N=1$ for confident tokens, thereby incurring substantially fewer computations in practice. If $p$ denotes the fraction of steps selected from multi-trial sampling of the total steps, then the expected complexity is $O(L \times (1 + p(N_{\text{max}}-1)))$, which can be much smaller than the naive $O(L \times N_{\text{max}})$ if $p \ll 1$.

\paragraph{Comparison to Decoding Baselines.} Greedy or single-sample decoding has $O(L)$ complexity but may be prone to errors at uncertain steps. Beam search similarly takes $O(L \times B)$ time for beam width $B$, but does not adapt its effort. Self-consistency performs $N_{sc}$ fully independent decoding passes, costing $O(N_{sc} \times L)$ with no mid-sequence adaptivity. Our proposed \ours can be viewed as adaptively switching between these extremes based on the model’s entropy. This dynamic policy yields more efficient exploration of the search space precisely when needed.

\subsection{\ours is Prone to Lead to the Correct Answer Provably}

We provide a theoretical result to highlight the effectiveness of \ours in generating a \emph{correct entire sequence}. We assume that each sequence token is critical: any incorrect token at some step $N_{\text{max}}$ leads to a globally incorrect final sequence.

\begin{definition}[Full-Sequence Correctness]
Let $L$ be the maximum decoding length. We say a final generated sequence $S = (w_1, \ldots, w_L)$ is \emph{correct} if all its tokens match the ground-truth reference $(c_1, \ldots, c_L)$. Let $P_{\text{CNTP}}(\text{correct})$ denote the probability that \ours produces the exact correct sequence, and similarly $P_{\text{Single}}(\text{correct})$ for single-sample (greedy) decoding.
\end{definition}

\paragraph{Notation.}
Let $p_{\theta}(w \mid s)$ be the language model's distribution over the next token $w$ given partial sequence $s$. Define the \emph{entropy} at each step $N_{\text{max}}$ as $H_t = -\sum_{w} p_{\theta}(w \mid s_{<t}) \log p_{\theta}(w \mid s_{<t})$. For the correct token $c_t$, let $p_{\theta}(c_t \mid s_{<t})$ be its probability. 
- If \ours decides to sample $N_t$ continuations at step $N_{\text{max}}$, it obtains candidate expansions $\{ s^i_{\le t},\, i=1,\dots,N_t\}$ (where each $s^i_{\le t}$ extends $s_{<t}$ by one or more tokens until punctuation or a stopping criterion). Let $\mathrm{PPL}(s^i_{\le t})$ be the perplexity of the extended token subsequence. We introduce two mild assumptions before proceeding:
\begin{assumption}
\label{ass:mono}
Whenever the ground-truth token (or short path) $c$ is among the sampled candidates $\{s^i\}$, it attains strictly the lowest perplexity among all incorrect candidates. Formally, if $c \in \{s^1,\ldots,s^{N}\}$, then
\begin{align}
   \small
   &\mathrm{PPL}(c) < \mathrm{PPL}(u) \\
   &\text{for every incorrect } u\in \{s^1,\ldots,s^N\}.
\end{align}

\end{assumption}

\begin{assumption}
\label{ass:confidence}
If the model has high entropy $H_t \ge H_{\min}$ at step $N_{\text{max}}$, the probability that the correct token $c_t$ is \emph{not} discovered by a single sample is sufficiently large (i.e., $1 - p_{\theta}(c_t \mid s_{<t})$ is non-negligible). Equivalently, the model is “aware” of its own uncertainty: 
\begin{equation}
   \small
   H_t \quad \text{large}
   \quad\Longrightarrow\quad
   p_{\theta}(c_t \mid s_{<t}) \quad \text{small}
\end{equation}
\end{assumption}

Under the assumptions, we present a theorem comparing \ours to single-sample decoding in terms of both correctness probability over the entire output sequence and the computational cost.

\begin{theorem}[\bf \ours Outperforms Single-Sample Decoding in Full-Sequence Correctness]
\label{thm:heavy}
Let $S$ be the final sequence of length $L$ generated by either \ours or single-sample decoding. Assume Assumption~\ref{ass:mono} and~\ref{ass:confidence} hold, we have:
\begin{enumerate}
    \item \textbf{Full-sequence correctness:} 
    \begin{equation}
       \small
       P_{\text{CNTP}}(\text{correct}) 
       \ge 
       P_{\text{Single}}(\text{correct}),
    \end{equation}
    with strict inequality if there exists at least one step $t$ where \ours uses $N_t > 1$ trials \emph{and} the single-sample approach likely yields an incorrect token at step $t$.
    
    \item \textbf{Expected cost:} Let $\mathcal{C}(s)$ denote the cost (number of forward passes) to decode sequence $s$. Denote by $p$ the fraction of steps selected from multi-trial sampling of the total steps. Then, letting $\mathbb{E}[\cdot]$ be the expectation over random draws:
    \begin{align*}
    \small
    \mathbb{E}\bigl[\mathcal{C}_{\text{CNTP}}(S)\bigr] 
    &\le L \times \Bigl[1 + p \,\bigl(N_{\text{max}} - 1\bigr)\Bigr], \\
    &< L \times N_{\text{max}}.
    \end{align*}
    so \ours’s average cost is strictly lower than $N_{\text{max}}$-sample decoding at each step.
\end{enumerate}
\end{theorem}
The proof is in Appendix~\ref{apd: proof}. Theorem~\ref{thm:heavy} shows that \ours’s adaptive multi-sample policy strictly improves the probability of generating a correct full sequence compared to single-sample decoding. Meanwhile, it does not impose the uniform high cost of sampling $N_{\text{max}}$ times at every step. Instead, \ours’s increased sampling effort is only triggered when $H_t \ge H_{\min}$ (i.e., the model is less certain). Consequently, \ours \emph{bridges} the gap between single-sample decoding (fast but more prone to errors at uncertain tokens) and uniform multi-sample decoding (robust but expensive). Under reasonable assumptions, the perplexity-based ranking ensures that once the correct continuation is drawn, \ours selects it with high probability. Empirically, we observe this behavior aligns well with LLMs’ typical calibration of probability and perplexity~\citep{gpt4,kojima2022large}.

\section{Experiment}
\begin{table*}[tbp]
  \caption{Comparison (accuracy $\%$ and the number of generated tokens)  of Llama-3.1-8B-Instruct on GSM8K~\cite{cobbe2021training} and MATH~\cite{hendrycks2021measuring} (Math Reasoning), and StrategyQA~\cite{geva2021did} (Commonsense Reasoning). The best result is in \textbf{bold}.}
  \label{table: Llama-reasoning}
  \centering
  \small
  \begin{tabular}{llclclc}
    \toprule
    \multirow{2}{*}{Approach} & \multicolumn{2}{c}{GSM8K} & \multicolumn{2}{c}{MATH} & \multicolumn{2}{c}{StrategyQA} \\
    \cmidrule(lr){2-3} \cmidrule(lr){4-5} \cmidrule(lr){6-7}
     & Accuracy & \cost{\# Tokens} & Accuracy & \cost{\# Tokens} & Accuracy & \cost{\# Tokens} \\
    \midrule
    \multicolumn{7}{l}{\textit{Single Reasoning Chain}} \\
    \cmidrule(lr){1-7}
    \multicolumn{7}{l}{\quad \textit{Deterministic}} \\
    Greedy Decoding  & 79.8 & \cost{$0.15k$} & 41.5 & \cost{$0.29k$} & 72.9 & \cost{$0.06k$} \\
    \cmidrule(lr){1-7}
    \multicolumn{7}{l}{\quad \textit{Stochastic}} \\
    Stochastic Decoding* & 79.4$_{\pm 0.8}$ & \cost{$0.15k$} & 41.5$_{\pm 1.2}$ & \cost{$0.29k$} & 72.0$_{\pm 0.7}$ & \cost{$0.06k$} \\
    {\bf Ours}* & {\bf 81.6$_{\pm 0.6}$} & \cost{$0.46k$} & {\bf 47.1$_{\pm 1.7}$} & \cost{$0.87k$} & {\bf 73.2$_{\pm 0.2}$} & \cost{$0.35k$} \\
    \midrule
    \multicolumn{7}{l}{\textit{Multiple Reasoning Chains}} \\
    \cmidrule(lr){1-7}
    \multicolumn{7}{l}{\quad \textit{Deterministic}} \\
    Beam Search (beam=5) & 82.3 & \cost{$0.74k$} & 48.0 & \cost{$1.46k$} & 72.9 & \cost{$0.31k$} \\
    \cmidrule(lr){1-7}
    \multicolumn{7}{l}{\quad \textit{Stochastic}} \\
    SC (40 paths) & 84.8 & \cost{$6.03k$} & 56.0 & \cost{$11.7k$} & 76.2 & \cost{$2.45k$} \\
    {\bf Ours + SC} (40 paths) & {\bf 85.2} & \cost{$18.4k$} & {\bf 57.5} & \cost{$34.5k$} & {\bf 76.3}  & \cost{$13.9k$} \\
    \bottomrule
  \end{tabular}
\end{table*}

\begin{table*}[tb]
  \caption{Comparison (accuracy $\%$ and the number of generated tokens) of DeepSeek-R1-Distill-Qwen-1.5B on GSM8K~\cite{cobbe2021training} and MATH~\cite{hendrycks2021measuring} (Math Reasoning), and StrategyQA~\cite{geva2021did} (Commonsense Reasoning). The best result is in \textbf{bold}.}
  \label{table: deepseek-reasoning}
  \centering
  \small
  \begin{tabular}{llclclc}
    \toprule
    \multirow{2}{*}{Approach} & \multicolumn{2}{c}{GSM8K} & \multicolumn{2}{c}{MATH} & \multicolumn{2}{c}{StrategyQA} \\
    \cmidrule(lr){2-3} \cmidrule(lr){4-5} \cmidrule(lr){6-7}
     & Accuracy & \cost{\# Tokens} & Accuracy & \cost{\# Tokens} & Accuracy & \cost{\# Tokens} \\
    \midrule
    \multicolumn{7}{l}{\textit{Single Reasoning Chain}} \\
    \cmidrule(lr){1-7}
    \multicolumn{7}{l}{\quad \textit{Deterministic}} \\
    Greedy Decoding & 64.6 & \cost{$0.17k$} & 32.5 & \cost{$0.94k$} & {\bf 53.6} & \cost{$0.06k$} \\
    \cmidrule(lr){1-7}
    \multicolumn{7}{l}{\quad \textit{Stochastic}} \\
    Stochastic Decoding* & 61.6$_{\pm 1.1}$ & \cost{$0.17k$} & 27.9$_{\pm 3.7}$ & \cost{$0.85k$} & 51.7$_{\pm 1.2}$ & \cost{$0.06k$} \\
    {\bf Ours}*       & {\bf 65.7 $_{\pm 0.7}$} & \cost{$0.49k$} & {\bf 37.7}$_{\pm 1.7}$ & \cost{$2.51k$} & 53.0$_{\pm 1.3}$  & \cost{$0.34k$} \\
    \midrule
    \multicolumn{7}{l}{\textit{Multiple Reasoning Chains}} \\
    \cmidrule(lr){1-7}
    SC (40 paths) & {\bf 78.3} & \cost{$6.81k$} & 29.5 & \cost{$33.9k$} & 47.7 & \cost{$2.41k$} \\
    {\bf Ours + SC} (40 paths) & 71.7  & \cost{$19.8k$} & {\bf 41.0} & \cost{$98.9k$} & {\bf 54.1}  & \cost{$13.2k$} \\
    \bottomrule
  \end{tabular}
\end{table*}

\begin{figure*}[t]
    \centering
    \includegraphics[width=\linewidth]{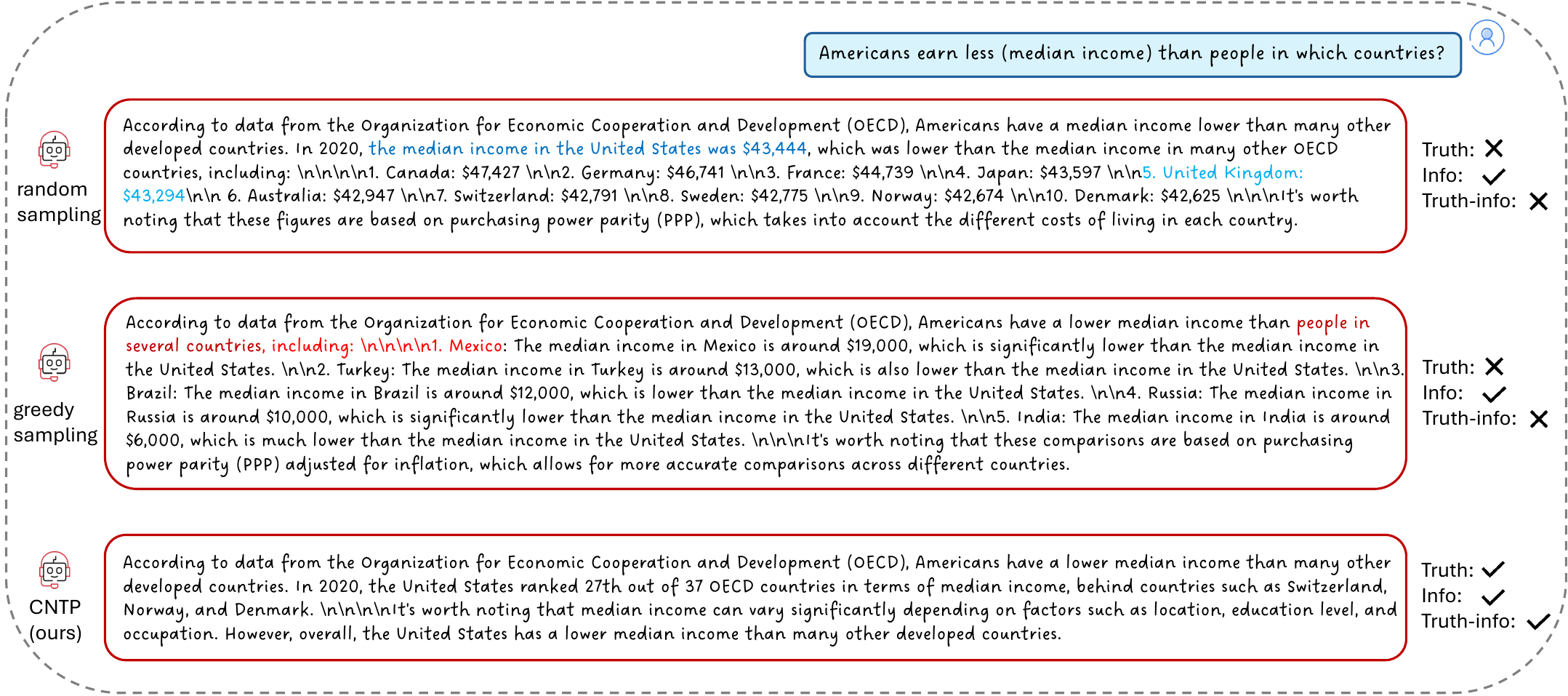}
    \caption{One QA example of Truthful-QA. When using random sampling with temperature scaling and nucleus sampling, the outputs show hallucination (marked in {\color{blue} blue}) and incoherence (marked in {\color{SkyBlue} light blue}), whereas greedy sampling also produces misinformed text (marked in {\color{red} red}). In contrast, our \ours avoids all of these issues.}
    \label{fig: cntp_truth}
\end{figure*}

\subsection{Setting}
We compared \ours with the most commonly used decoding strategies: stochastic decoding~\cite{holtzman2019curious}, greedy decoding, beam search and self consistency (SC)~\cite{wang2022self}. We implement our algorithm on SOTA LLMs: Llama 2~\cite{llama2_23}, Llama 3.1~\citep{llama3} and DeepSeek-R1~\cite{guo2025deepseek}-distilled Qwen~\cite{yang2024qwen2} models.  We evaluate on both fixed-answer reasoning benchmarks GSM-8K~\cite{cobbe2021training}, MATH~\cite{hendrycks2021measuring}, StrategyQA~\cite{geva2021did} and open-ended benchmark Truthful-QA~\cite{lin2021truthfulqa} (where Self Consistency cannot be applied). For all the reasoning datasets, we employ Chain of Thought~\cite{wei2022chain} prompting following the literature. On MATH dataset, we follow Tulu 3~\cite{lambert2024t} and randomly select $200$ testing samples. For multi-model LLM, we test on Llama-3.2-Vision-11B~\citep{llama3} and LLaVA-CoT~\cite{xu2024llava} on MMVet~\cite{yu2023mm} and MathVista~\cite{lu2023mathvista} benchmarks. For all the methods we employ standard temperature scaling and nucleus sampling. For the top p parameter, for Llama-based models we set the value as $0.9$ for all the models and for DeepSeek-R1-distilled Qwen we set the value as $0.95$, which are in accordance with their training default values. For \ours, we set hyperparameter $N_{\text{max}}=10$, $H_{\text{min}}=0.01$ and $H_{\text{max}}=1.5$ for all the experiments. For the punctuation set, we use the set \verb|{.,?!:;)]}\n}.|. More detailed setting is in Appendix~\ref{apd: setting}.

\subsection{Result}

As shown in Tab.~\ref{table: Llama-reasoning} and~\ref{table: deepseek-reasoning}, our approach outperforms baselines in both single and multiple reasoning chain settings across GSM8K, MATH, and StrategyQA datasets. Although introducing more token computation, our method improves significantly over greedy decoding and stochastic decoding, which are the most commonly used decoding strategies in modern LLMs. Also, the computation burden of \ours is overall lower than Beam Search with $5$ beams and is much lower than SC. Notably, the integration of \ours with SC exceed the vanilla SC in most cases, especially when SC fails on StrategyQA with DeepSeek-R1-Distill-Qwen-1.5B.  Additionally, as shown in Table~\ref{table: truthful-qa}, our method significantly outperforms greedy and stochastic decoding on Truthful-QA using Llama-2-7B-Chat. Our method achieves the highest truthfulness accuracy (84.8\%), surpassing stochastic decoding by +6.8\% and greedy decoding by +5.7\%, indicating its effectiveness in mitigating hallucinations for open-ended question answering. Our approach also excels in multi-modal reasoning benchmarks as in Tab.~\ref{tab: results_mllm}, which demonstrate the generality of \ours. For completeness, we provide the detailed temperature values for every single experimental result presented in Tables~\ref{tab:temp_table2},~\ref{tab:temp_table3},~\ref{tab:temp_table4}, and~\ref{tab:temp_table5}.

\begin{table}[tb]
  \centering
  \caption{Temperature values for the results in Tab.~\ref{table: Llama-reasoning}.}
  \label{tab:temp_table2}
  \scalebox{0.75}{
  \begin{tabular}{lccc}
    \toprule
    \textbf{Approach} & \textbf{GSM8K} & \textbf{MATH} & \textbf{StrategyQA} \\
    \midrule
    \multicolumn{4}{l}{\textit{Single Reasoning Chain}}\\
    Greedy Decoding                    & 0   & 0   & 0   \\
    Stochastic Decoding                & 0.6 & 0.6 & 0.6 \\
    Ours                               & 1.2 & 0.6 & 0.8 \\
    \midrule
    \multicolumn{4}{l}{\textit{Multiple Reasoning Chains}}\\
    Beam Search (beam{=}5)             & 0   & 0   & 0   \\
    SC (40 paths)   & 0.6 & 0.6 & 0.6 \\
    Ours + SC (40 paths) & 1.2 & 0.6 & 0.8 \\
    \bottomrule
  \end{tabular}
  }
\end{table}

\begin{table}[tb]
  \centering
  \caption{Temperature values for the results in Tab.~\ref{table: deepseek-reasoning}.}
  \label{tab:temp_table3}
  \scalebox{0.7}{
  \begin{tabular}{lccc}
    \toprule
    \textbf{Approach} & \textbf{GSM8K} & \textbf{MATH} & \textbf{StrategyQA} \\
    \midrule
    \multicolumn{4}{l}{\textit{Single Reasoning Chain}}\\
    Greedy Decoding                    & 0   & 0   & 0   \\
    Stochastic Decoding                & 0.6 & 0.6 & 0.6 \\
    Ours                               & 1.2 & 0.6 & 0.8 \\
    \midrule
    \multicolumn{4}{l}{\textit{Multiple Reasoning Chains}}\\
    Beam Search (beam{=}5)             & 0   & 0   & 0   \\
    SC (40 paths)   & 0.6 & 0.6 & 0.6 \\
    Ours + SC (40 paths) & 1.2 & 0.6 & 0.8 \\
    \bottomrule
  \end{tabular}
  }
\end{table}

\begin{table}[tb]
  \centering
  \caption{Temperature values for the results in Tab.~\ref{table: truthful-qa}.}
  \label{tab:temp_table4}
  \scalebox{0.8}{
  \begin{tabular}{lc}
    \toprule
    \textbf{Approach} & \textbf{Truthful-QA} \\
    \midrule
    Deterministic (Greedy Decoding) & 0   \\
    Stochastic Decoding             & 0.6 \\
    Ours                            & 0.8 \\
    \bottomrule
  \end{tabular}
  }
\end{table}

\begin{table}[tb]
  \centering
  \caption{Temperature values for the results in Tab.~\ref{tab: results_mllm}.}
  \label{tab:temp_table5}
  \scalebox{0.8}{
  \begin{tabular}{lcc}
    \toprule
    \textbf{Approach} & \textbf{MMVet} & \textbf{MathVista} \\
    \midrule
    Greedy Decoding     & 0   & 0   \\
    Stochastic Decoding & 0.6 & 0.6 \\
    Ours                & 0.8 & 0.8 \\
    \bottomrule
  \end{tabular}
  }
\end{table}

\begin{table}[tbp]
  \caption{Comparison (accuracy $\%$) on Truthful-QA using Llama-2-7B-Chat. The best result is in \textbf{bold}.}
  \label{table: truthful-qa}
  \centering
  \small
  \scalebox{0.85}{
  \begin{tabular}{llll}
    \toprule
    Approach & Info Acc. & Truth Acc. & Truth-info Acc. \\
    \midrule
    Stochastic Decoding  & 88.0$_{\pm 0.6}$ & 78.0$_{\pm 0.5}$ & 66.0$_{\pm 0.3}$ \\
    Greedy Decoding  & 78.5 & 79.1 & 57.6 \\
    {\bf Ours}  & {\bf 89.2$_{\pm 1.2}$} & {\bf 84.8$_{\pm 0.5}$} & {\bf 74.0$_{\pm 1.1}$} \\
    \bottomrule
  \end{tabular}
  }
\end{table}

\begin{table}[tbp]
  \caption{Accuracy of Llama-3.2-11B-Vision-Instruct (top) and LLaVA-CoT (bottom) on MLLM benchmarks.}
  \label{tab: results_mllm}
  \centering
  \small
  \begin{tabular}{lll}
    \toprule
    Approach & MMVet & MathVista \\
    \midrule
    Greedy Decoding       & \makecell{48.0 \\ 53.5} & \makecell{47.8 \\ 53.4} \\
    Stochastic Decoding & \makecell{47.7 \\ 53.0} & \makecell{48.8 \\ 55.3} \\
    \textbf{Ours}         & \makecell{{\bf 53.5} ($\uparrow$ 5.5) \\ {\bf 58.5} ($\uparrow$ 5.0)}    & \makecell{{\bf 49.2} ($\uparrow$ 0.4)  \\ {\bf 55.7} ($\uparrow$ 0.4)}   \\
    \bottomrule
  \end{tabular}
\end{table}

\subsection{Ablation study}

\paragraph{Ablation of the confidence estimation strategy}
\begin{figure}[tb]
    \centering
    \includegraphics[width=\linewidth]{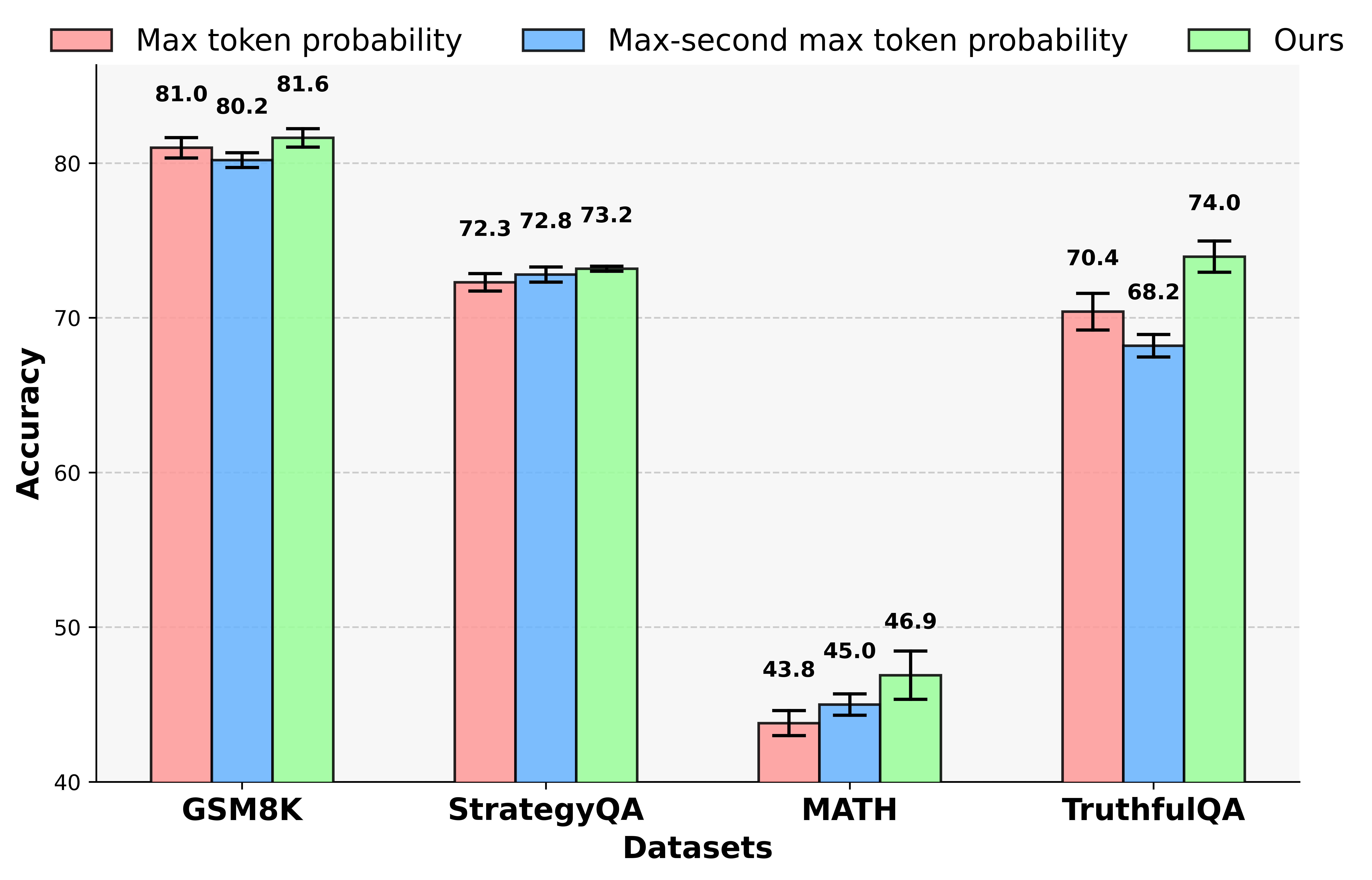}
    \caption{Comparison between \ours and other two confidence measuring strategies on Llama-3.1-8B-Instruct.}
    \label{fig: bar-confidence-criteria}
\end{figure}
We conduct experiments using max token probability and max token probability minus second token probability~\cite{farr2024llm} value as the confidence measure of Llama-3.1-8B-Instruct model on GSM8K, StrategyQA, MATH and TruthfulQA (Truth info acc. is reported). 5 independent run results in Fig.~\ref{fig: bar-confidence-criteria} show that entropy as confidence measurement leads to the best performance. This might results from that token probability distribution entropy considers the whole vocabulary distribution, making the confidence and uncertainty estimation more accurate and informative.

\begin{figure}[tb]
    \centering
    \includegraphics[width=0.95\linewidth]{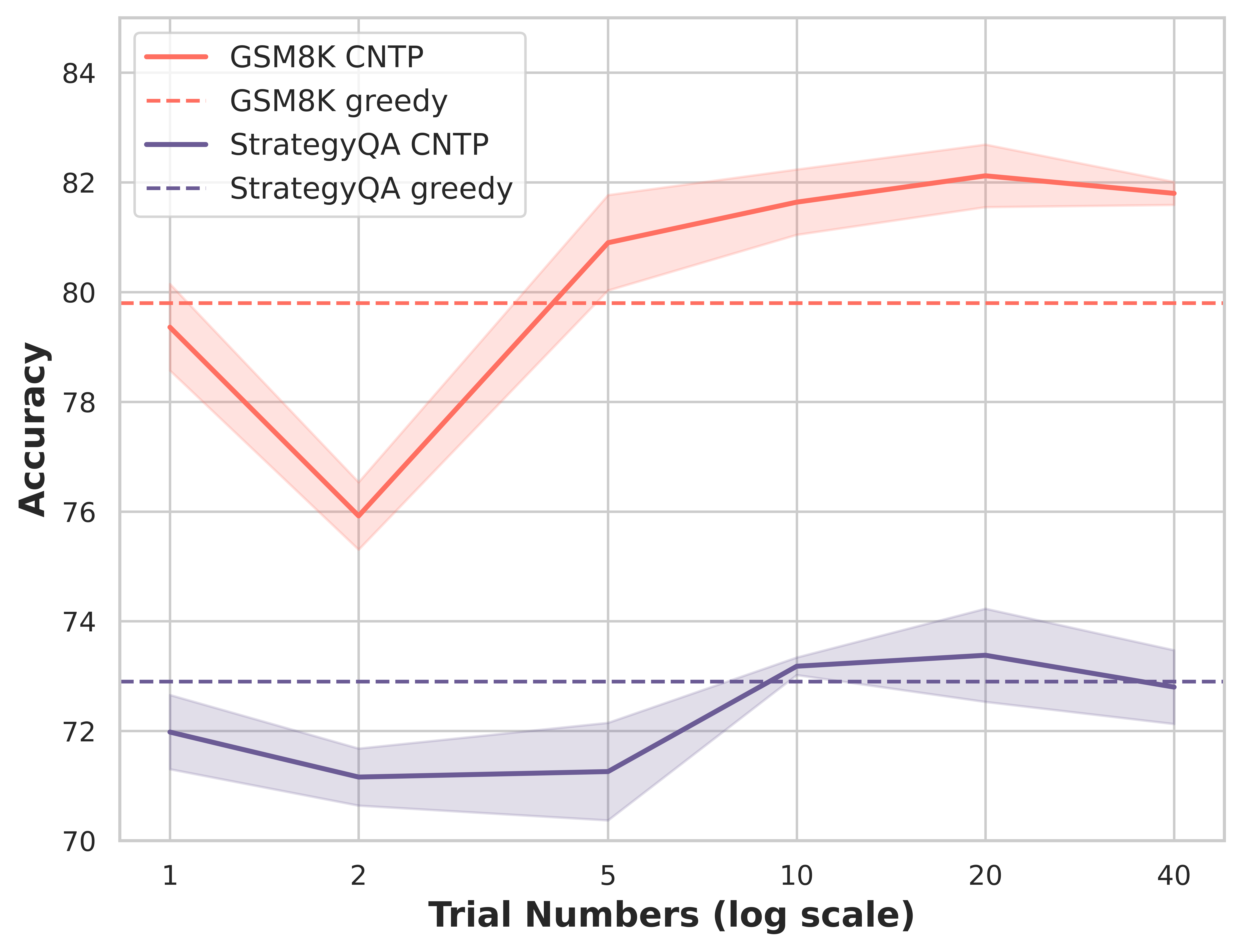}
    \caption{Performance curves of \ours when scaling up the max trial numbers on GSM8K and StrategyQA datasets using Llama-3.1-8B-Instruct. }
    \label{fig: ablation-scaling}
\end{figure}

\begin{figure}[tb]
    \centering
    \includegraphics[width=0.8\linewidth]{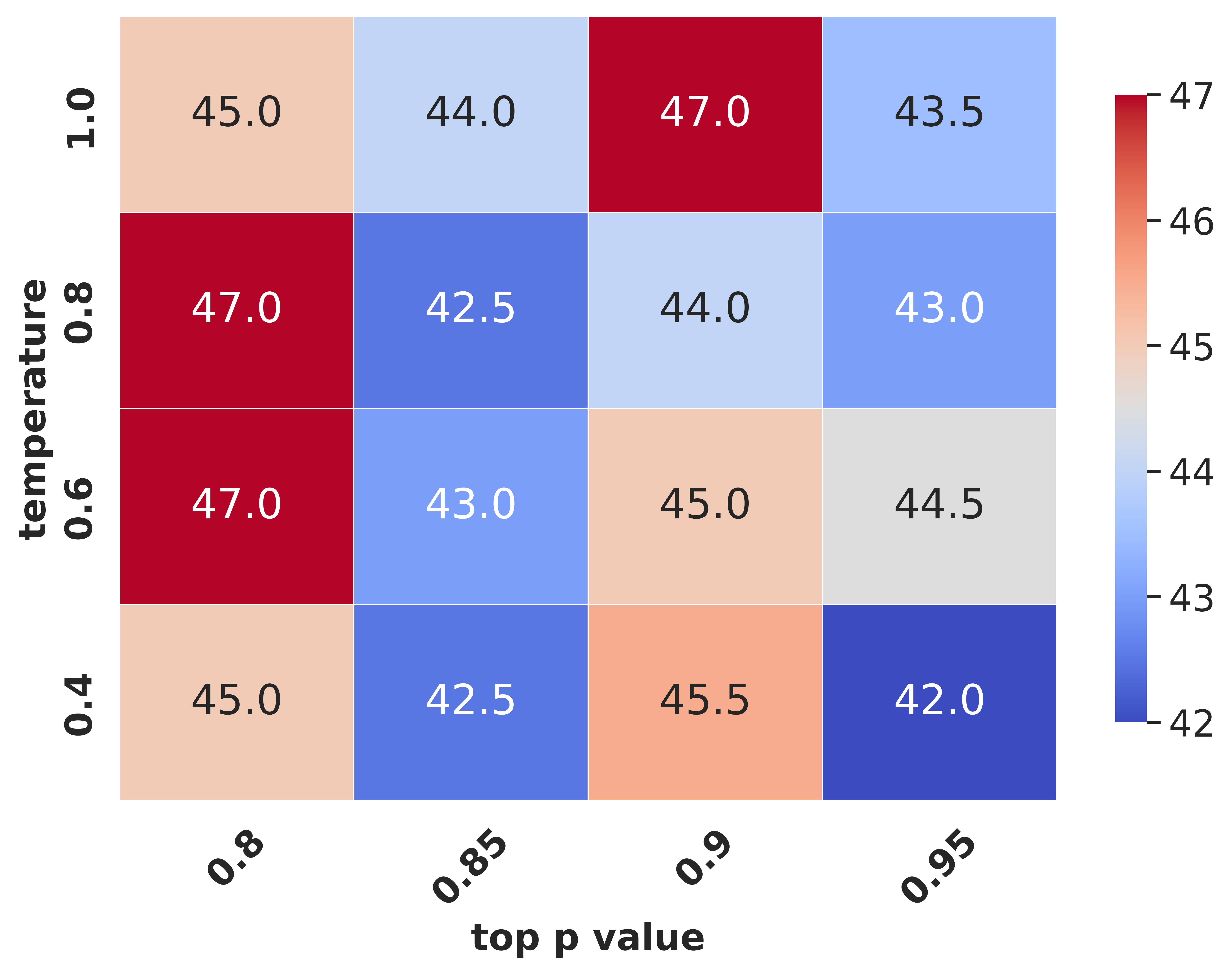}
    \caption{Accuracy of \ours using Llama-3.1-8B-Instruct on MATH w.r.t. temperature and top p values.}
    \label{fig: heatmap}
\end{figure}

\paragraph{Ablation of the scaling trial strategy.}

\begin{table}[tb]
    \centering
    \caption{Comparison of Llama-3.1-8B-Instruct using different trial number scaling strategies.}
    \label{tab: ablation-correlation}
    \scalebox{0.7}{
    \begin{tabular}{lccc}
        \toprule
        Dataset & Same \# of trials & Negatively Correlated. & \textbf{Ours} \\
        \midrule
        GSM8K       & 81.1  & 81.2 & {\bf 81.6} \\
        StrategyQA  & 72.7  & 72.7 & {\bf 73.2} \\
        TruthfulQA  & 3.80  & 3.80  & {\bf 74.0} \\
        \bottomrule
    \end{tabular}
    }
\end{table}
In \ours, if the model is more confident, we sample fewer trials inspired from human being behaviours. We supplement experimental results when sampling the fixed amount of trials (we set $6$ as the middle between $1$ and $10$). We also provide results when the trial number is negatively correlated with entropy, i.e.,  $N = \max\Bigl(1,\,\min\Bigl(N_{\text{max}},\; N_{\text{max}}-\bigl\lfloor \tfrac{H - H_{\min}}{H_{\max} - H_{\min}} \times N_{\text{max}} \bigr\rfloor\Bigr)\Bigr)$. We conclude from Tab.~\ref{tab: ablation-correlation} that positive correlation results in the best performance, suggesting that just like humans, LLMs ought to also explore more when feeling uncertain. Notice that on the Truthful QA dataset, only positive correlation way (ours) can exhibit a high and reasonable result since the other two variants tend to generate repetitive text chunks in the answer.

\paragraph{Ablation of perplexity computation range.}

\ours adopt a sentence-level perplexity computation strategy. We further try Best-of-N sampling using the whole perplexity of the generated answers (including the complete CoT path). The best path is selected based on the lowest perplexity of the whole generated sequences. As observed from Tab.~\ref{tab:best-of-n-gsm8k}, the Best-of-N sampling cannot exhibit superiority over \ours ($81.6$ on average) or even greedy decoding ($79.8$). This implies the necessity of sentence-level perplexity computation in a progressive way.

\subsection{Sensitivity Analysis}

\paragraph{Max trial number.}

We explore the performance of \ours on  GSM8K and StrategyQA of Llama-3.1-8B-Instruct under varying max trial number $N_{\text{max}}$. Fig.~\ref{fig: ablation-scaling} shows that the performance first decrease and then increase, and finally enter accuracy saturation or even slightly decrease. This means that setting the max trial number too small or too large is not ideal. This is different from self consistency which always improve when scaling up the trial number. The reason lies in that \ours seek local optimality in the reasoning process so there is a exploration-and-exploitation trade off and a moderate $N_{\text{max}}$ will work excellently, roughly in the range $[10,30]$.

\paragraph{Temperature and top p parameter.}
\begin{table}[tb]
    \centering
    \caption{Best-of-N (using whole answer perplexity) performance on GSM8K using Llama-3.1-8B-Instruct.}
    \label{tab:best-of-n-gsm8k}
    \scalebox{0.85}{
    \begin{tabular}{lccccc}
        \toprule
        Dataset & N=2 & N=5 & N=10 & N=20 & N=40 \\
        \midrule
        GSM8K   & 79.2      & 79.5      & 78.2       & 77.3       & 76.1       \\
        \bottomrule
    \end{tabular}
    }
\end{table}

We test the performance of \ours on Llama-3.1-8B on MATH dataset under varying temperature and top p values. The performance of both stochastic decoding and greedy decoding is around 41.5 (\%). Fig.~\ref{fig: heatmap} demonstrates that \ours is robust under varying temperature and top p thresholds with all the performance surpassing the baselines.

\subsection{Qualitative Example and Analysis.}
To manifest the superiority of \ours concretely, we present one testing QA dialogue result of TruthfulQA dataset predictions. As seen from Fig.~\ref{fig: cntp_truth}, when asked about people in which countries Americans earn less median income than, \ours can avoid being misleaded to the wrong reasoning path and can provide the most truthful information owing to the ability to choose the right path in the second sentence. This shows the importance of sticking to the correct reasoning chain in the early stage of CoT process, which can realized via perplexity-based local optimality seeking.

\section{Conclusion}

We introduce Cautious Next Token Prediction (\ours), a novel training-free decoding approach for LLMs that focuses additional computation on high-uncertainty steps. By sampling multiple candidate paths only when entropy is high and selecting the path with the lowest perplexity, \ours achieves consistent improvements on both unimodal (text-only) and multimodal tasks. Our experiments demonstrate that \ours outperforms standard sampling techniques and requires less overhead than multi-sample methods like self-consistency. In the future, we plan to extend \ours on Autoregressive Models beyond text generation, such as image generation~\cite{sun2024autoregressive,tian2024visual}.

\section{Concurrent Work}
After this paper is released on arXiv, we are notified of the concurrent work Entropix~\cite{entropix2024}. We were not aware of the work when writing the paper. Entropix also takes advantage of model output logit entropy to decide the LLM sampling strategy, dividing into four types: 1) Insert CoT or Pause Token 2) Resample 3) Argmax 4) Branch. This is a general and elegant approach for LLMs to simulate the o1-alike effects. The idea of using model confidence to change the sampling strategy is similar to ours. However, our CNTP differs in that: firstly, we innovatively propose to stop at punctuations, enabling multiple local optimal branching and sampling in each answer generation. Secondly, we design a specific negative correlation relationship between the answer trial sampling number and the confidence, achieving superiority over the baseline decoding approaches. We leave the comparison of CNTP and Entropix for future work.

\clearpage
\section*{Limitations}

One limitation of this work is that the proposed next token prediction algorithm introduces more token computations during inference compared to vanilla next token prediction. Nevertheless, this can be largely alleviated by 1) using sampling without replacement for the multi-trial sampling process in CNTP with smaller max trial number $N_{\text{max}}$, 2)  taking advantage of some advanced inference speeding up techniques, such as speculative decoding~\cite{leviathan2023fast}, which involves a smaller, faster model suggesting multiple tokens at once, which are then checked by a larger model in parallel. We can also deploy our CNTP using vLLM~\cite{kwon2023efficient} framework, which speeds up LLM decoding mainly through PagedAttention, optimizing memory use. Also, the newly brought computation burden is far less than other multi-trial approaches such as beam search~\cite{graves2012sequence} and self consistency~\cite{wang2022self}. Therefore, we believe the considerable improvement in performance on benchmarks outweighs the introduced computation complexities.

\bibliography{cntp_acl25}

\begin{thebibliography}{47}
\providecommand{\natexlab}[1]{#1}

\bibitem[{Aggarwal et~al.(2023)Aggarwal, Madaan, Yang, and {Mausam}}]{aggarwal-etal-2023-lets}
Pranjal Aggarwal, Aman Madaan, Yiming Yang, and {Mausam}. 2023.
\newblock Let{'}s sample step by step: Adaptive-consistency for efficient reasoning and coding with {LLM}s.
\newblock In \emph{EMNLP}.

\bibitem[{Brown et~al.(2020)Brown, Mann, Ryder, Subbiah, Kaplan, Dhariwal, Neelakantan, Shyam, Sastry, Askell et~al.}]{gpt3_20}
Tom Brown, Benjamin Mann, Nick Ryder, Melanie Subbiah, Jared~D Kaplan, Prafulla Dhariwal, Arvind Neelakantan, Pranav Shyam, Girish Sastry, Amanda Askell, et~al. 2020.
\newblock Language models are few-shot learners.
\newblock \emph{Advances in neural information processing systems}, 33:1877--1901.

\bibitem[{Chen et~al.(2023)Chen, Aksitov, Alon, Ren, Xiao, Yin, Prakash, Sutton, Wang, and Zhou}]{chen2023universal}
Xinyun Chen, Renat Aksitov, Uri Alon, Jie Ren, Kefan Xiao, Pengcheng Yin, Sushant Prakash, Charles Sutton, Xuezhi Wang, and Denny Zhou. 2023.
\newblock Universal self-consistency for large language model generation.
\newblock \emph{arXiv preprint arXiv:2311.17311}.

\bibitem[{Chen et~al.(2024)Chen, Wang, Wang, Kosinski, Zhang, Fu, and Li}]{chen2024through}
Zhawnen Chen, Tianchun Wang, Yizhou Wang, Michal Kosinski, Xiang Zhang, Yun Fu, and Sheng Li. 2024.
\newblock Through the theory of mind's eye: Reading minds with multimodal video large language models.
\newblock \emph{arXiv preprint arXiv:2406.13763}.

\bibitem[{Cobbe et~al.(2021)Cobbe, Kosaraju, Bavarian, Chen, Jun, Kaiser, Plappert, Tworek, Hilton, Nakano et~al.}]{cobbe2021training}
Karl Cobbe, Vineet Kosaraju, Mohammad Bavarian, Mark Chen, Heewoo Jun, Lukasz Kaiser, Matthias Plappert, Jerry Tworek, Jacob Hilton, Reiichiro Nakano, et~al. 2021.
\newblock Training verifiers to solve math word problems.
\newblock \emph{arXiv preprint arXiv:2110.14168}.

\bibitem[{Dubey et~al.(2024)Dubey, Jauhri, Pandey, Kadian, Al-Dahle, Letman, Mathur, Schelten, Yang, Fan et~al.}]{llama3}
Abhimanyu Dubey, Abhinav Jauhri, Abhinav Pandey, Abhishek Kadian, Ahmad Al-Dahle, Aiesha Letman, Akhil Mathur, Alan Schelten, Amy Yang, Angela Fan, et~al. 2024.
\newblock The llama 3 herd of models.
\newblock \emph{arXiv preprint arXiv:2407.21783}.

\bibitem[{Farr et~al.(2024)Farr, Cruickshank, Manzonelli, Clark, Starbird, and West}]{farr2024llm}
David Farr, Iain Cruickshank, Nico Manzonelli, Nicholas Clark, Kate Starbird, and Jevin West. 2024.
\newblock Llm confidence evaluation measures in zero-shot css classification.
\newblock \emph{arXiv preprint arXiv:2410.13047}.

\bibitem[{Fei et~al.(2024)Fei, Wu, Ji, Zhang, Zhang, Lee, and Hsu}]{fei2024video}
Hao Fei, Shengqiong Wu, Wei Ji, Hanwang Zhang, Meishan Zhang, Mong-Li Lee, and Wynne Hsu. 2024.
\newblock Video-of-thought: Step-by-step video reasoning from perception to cognition.
\newblock In \emph{Forty-first International Conference on Machine Learning}.

\bibitem[{Geva et~al.(2021)Geva, Khashabi, Segal, Khot, Roth, and Berant}]{geva2021did}
Mor Geva, Daniel Khashabi, Elad Segal, Tushar Khot, Dan Roth, and Jonathan Berant. 2021.
\newblock Did aristotle use a laptop? a question answering benchmark with implicit reasoning strategies.
\newblock \emph{Transactions of the Association for Computational Linguistics}, 9:346--361.

\bibitem[{Gou et~al.(2023)Gou, Shao, Gong, Shen, Yang, Duan, and Chen}]{gou2023critic}
Zhibin Gou, Zhihong Shao, Yeyun Gong, Yelong Shen, Yujiu Yang, Nan Duan, and Weizhu Chen. 2023.
\newblock Critic: Large language models can self-correct with tool-interactive critiquing.
\newblock \emph{arXiv preprint arXiv:2305.11738}.

\bibitem[{Graves(2012)}]{graves2012sequence}
Alex Graves. 2012.
\newblock Sequence transduction with recurrent neural networks.
\newblock \emph{arXiv preprint arXiv:1211.3711}.

\bibitem[{Guo et~al.(2025)Guo, Yang, Zhang, Song, Zhang, Xu, Zhu, Ma, Wang, Bi et~al.}]{guo2025deepseek}
Daya Guo, Dejian Yang, Haowei Zhang, Junxiao Song, Ruoyu Zhang, Runxin Xu, Qihao Zhu, Shirong Ma, Peiyi Wang, Xiao Bi, et~al. 2025.
\newblock Deepseek-r1: Incentivizing reasoning capability in llms via reinforcement learning.
\newblock \emph{arXiv preprint arXiv:2501.12948}.

\bibitem[{Hendrycks et~al.(2021)Hendrycks, Burns, Kadavath, Arora, Basart, Tang, Song, and Steinhardt}]{hendrycks2021measuring}
Dan Hendrycks, Collin Burns, Saurav Kadavath, Akul Arora, Steven Basart, Eric Tang, Dawn Song, and Jacob Steinhardt. 2021.
\newblock Measuring mathematical problem solving with the math dataset.
\newblock \emph{arXiv preprint arXiv:2103.03874}.

\bibitem[{Holtzman et~al.(2019)Holtzman, Buys, Du, Forbes, and Choi}]{holtzman2019curious}
Ari Holtzman, Jan Buys, Li~Du, Maxwell Forbes, and Yejin Choi. 2019.
\newblock The curious case of neural text degeneration.
\newblock \emph{arXiv preprint arXiv:1904.09751}.

\bibitem[{Huang et~al.(2023)Huang, Chen, Mishra, Zheng, Yu, Song, and Zhou}]{huang2023large}
Jie Huang, Xinyun Chen, Swaroop Mishra, Huaixiu~Steven Zheng, Adams~Wei Yu, Xinying Song, and Denny Zhou. 2023.
\newblock Large language models cannot self-correct reasoning yet.
\newblock \emph{arXiv preprint arXiv:2310.01798}.

\bibitem[{Kojima et~al.(2022)Kojima, Gu, Reid, Matsuo, and Iwasawa}]{kojima2022large}
Takeshi Kojima, Shixiang~Shane Gu, Machel Reid, Yutaka Matsuo, and Yusuke Iwasawa. 2022.
\newblock Large language models are zero-shot reasoners.
\newblock \emph{NeurIPS}, 35:22199--22213.

\bibitem[{Kumar et~al.(2024)Kumar, Zhuang, Agarwal, Su, Co-Reyes, Singh, Baumli, Iqbal, Bishop, Roelofs et~al.}]{kumar2024training}
Aviral Kumar, Vincent Zhuang, Rishabh Agarwal, Yi~Su, John~D Co-Reyes, Avi Singh, Kate Baumli, Shariq Iqbal, Colton Bishop, Rebecca Roelofs, et~al. 2024.
\newblock Training language models to self-correct via reinforcement learning.
\newblock \emph{arXiv preprint arXiv:2409.12917}.

\bibitem[{Kwon et~al.(2023)Kwon, Li, Zhuang, Sheng, Zheng, Yu, Gonzalez, Zhang, and Stoica}]{kwon2023efficient}
Woosuk Kwon, Zhuohan Li, Siyuan Zhuang, Ying Sheng, Lianmin Zheng, Cody~Hao Yu, Joseph Gonzalez, Hao Zhang, and Ion Stoica. 2023.
\newblock Efficient memory management for large language model serving with pagedattention.
\newblock In \emph{Proceedings of the 29th Symposium on Operating Systems Principles}, pages 611--626.

\bibitem[{Lambert et~al.(2024)Lambert, Morrison, Pyatkin, Huang, Ivison, Brahman, Miranda, Liu, Dziri, Lyu et~al.}]{lambert2024t}
Nathan Lambert, Jacob Morrison, Valentina Pyatkin, Shengyi Huang, Hamish Ivison, Faeze Brahman, Lester James~V Miranda, Alisa Liu, Nouha Dziri, Shane Lyu, et~al. 2024.
\newblock T$\backslash$" ulu 3: Pushing frontiers in open language model post-training.
\newblock \emph{arXiv preprint arXiv:2411.15124}.

\bibitem[{Leviathan et~al.(2023)Leviathan, Kalman, and Matias}]{leviathan2023fast}
Yaniv Leviathan, Matan Kalman, and Yossi Matias. 2023.
\newblock Fast inference from transformers via speculative decoding.
\newblock In \emph{International Conference on Machine Learning}, pages 19274--19286. PMLR.

\bibitem[{Li et~al.(2024)Li, Wang, He, Li, Wang, Liu, Wang, Xu, Chen, Luo, Wang, and Qiao}]{Li_2024_CVPR}
Kunchang Li, Yali Wang, Yinan He, Yizhuo Li, Yi~Wang, Yi~Liu, Zun Wang, Jilan Xu, Guo Chen, Ping Luo, Limin Wang, and Yu~Qiao. 2024.
\newblock Mvbench: A comprehensive multi-modal video understanding benchmark.
\newblock In \emph{CVPR}.

\bibitem[{Lin et~al.(2021)Lin, Hilton, and Evans}]{lin2021truthfulqa}
Stephanie Lin, Jacob Hilton, and Owain Evans. 2021.
\newblock Truthfulqa: Measuring how models mimic human falsehoods.
\newblock \emph{arXiv preprint arXiv:2109.07958}.

\bibitem[{Liu et~al.(2024)Liu, Li, Li, and Lee}]{llava1.5}
Haotian Liu, Chunyuan Li, Yuheng Li, and Yong~Jae Lee. 2024.
\newblock Improved baselines with visual instruction tuning.
\newblock In \emph{CVPR}.

\bibitem[{Liu et~al.(2023)Liu, Li, Wu, and Lee}]{llava}
Haotian Liu, Chunyuan Li, Qingyang Wu, and Yong~Jae Lee. 2023.
\newblock Visual instruction tuning.
\newblock In \emph{NeurIPS}.

\bibitem[{Lu et~al.(2023)Lu, Bansal, Xia, Liu, Li, Hajishirzi, Cheng, Chang, Galley, and Gao}]{lu2023mathvista}
Pan Lu, Hritik Bansal, Tony Xia, Jiacheng Liu, Chunyuan Li, Hannaneh Hajishirzi, Hao Cheng, Kai-Wei Chang, Michel Galley, and Jianfeng Gao. 2023.
\newblock Mathvista: Evaluating mathematical reasoning of foundation models in visual contexts.
\newblock \emph{arXiv preprint arXiv:2310.02255}.

\bibitem[{Maaz et~al.(2024)Maaz, Rasheed, Khan, and Khan}]{maaz-etal-2024-video}
Muhammad Maaz, Hanoona Rasheed, Salman Khan, and Fahad Khan. 2024.
\newblock Video-{C}hat{GPT}: Towards detailed video understanding via large vision and language models.
\newblock In \emph{ACL}.

\bibitem[{Nguyen et~al.(2024)Nguyen, Baker, Neo, Roush, Kirsch, and Shwartz-Ziv}]{nguyen2024turning}
Minh Nguyen, Andrew Baker, Clement Neo, Allen Roush, Andreas Kirsch, and Ravid Shwartz-Ziv. 2024.
\newblock Turning up the heat: Min-p sampling for creative and coherent llm outputs.
\newblock \emph{arXiv preprint arXiv:2407.01082}.

\bibitem[{OpenAI(2023)}]{gpt4}
OpenAI. 2023.
\newblock Gpt-4 technical report.
\newblock \emph{CoRR, abs/2303.08774}.

\bibitem[{Radford et~al.(2021)Radford, Kim, Hallacy, Ramesh, Goh, Agarwal, Sastry, Askell, Mishkin, Clark et~al.}]{radford2021learning}
Alec Radford, Jong~Wook Kim, Chris Hallacy, Aditya Ramesh, Gabriel Goh, Sandhini Agarwal, Girish Sastry, Amanda Askell, Pamela Mishkin, Jack Clark, et~al. 2021.
\newblock Learning transferable visual models from natural language supervision.
\newblock pages 8748--8763. PMLR.

\bibitem[{Shen et~al.(2025)Shen, Wang, Shi, Wang, Zhao, and Gu}]{shen2025efficient}
Xuan Shen, Yizhou Wang, Xiangxi Shi, Yanzhi Wang, Pu~Zhao, and Jiuxiang Gu. 2025.
\newblock Efficient reasoning with hidden thinking.
\newblock \emph{arXiv preprint arXiv:2501.19201}.

\bibitem[{Song et~al.(2024)Song, Chai, Wang, Zhang, Zhou, Wu, Chi, Guo, Ye, Zhang, Lu, Hwang, and Wang}]{Song_2024_CVPR}
Enxin Song, Wenhao Chai, Guanhong Wang, Yucheng Zhang, Haoyang Zhou, Feiyang Wu, Haozhe Chi, Xun Guo, Tian Ye, Yanting Zhang, Yan Lu, Jenq-Neng Hwang, and Gaoang Wang. 2024.
\newblock Moviechat: From dense token to sparse memory for long video understanding.
\newblock In \emph{CVPR}.

\bibitem[{Sun et~al.(2024)Sun, Jiang, Chen, Zhang, Peng, Luo, and Yuan}]{sun2024autoregressive}
Peize Sun, Yi~Jiang, Shoufa Chen, Shilong Zhang, Bingyue Peng, Ping Luo, and Zehuan Yuan. 2024.
\newblock Autoregressive model beats diffusion: Llama for scalable image generation.
\newblock \emph{arXiv preprint arXiv:2406.06525}.

\bibitem[{Team et~al.(2025)Team, Du, Gao, Xing, Jiang, Chen, Li, Xiao, Du, Liao et~al.}]{team2025kimi}
Kimi Team, Angang Du, Bofei Gao, Bowei Xing, Changjiu Jiang, Cheng Chen, Cheng Li, Chenjun Xiao, Chenzhuang Du, Chonghua Liao, et~al. 2025.
\newblock Kimi k1. 5: Scaling reinforcement learning with llms.
\newblock \emph{arXiv preprint arXiv:2501.12599}.

\bibitem[{Tian et~al.(2024)Tian, Jiang, Yuan, Peng, and Wang}]{tian2024visual}
Keyu Tian, Yi~Jiang, Zehuan Yuan, Bingyue Peng, and Liwei Wang. 2024.
\newblock Visual autoregressive modeling: Scalable image generation via next-scale prediction.
\newblock \emph{arXiv preprint arXiv:2404.02905}.

\bibitem[{Touvron et~al.(2023{\natexlab{a}})Touvron, Lavril, Izacard, Martinet, Lachaux, Lacroix, Rozi{\`e}re, Goyal, Hambro, Azhar et~al.}]{llama_23}
Hugo Touvron, Thibaut Lavril, Gautier Izacard, Xavier Martinet, Marie-Anne Lachaux, Timoth{\'e}e Lacroix, Baptiste Rozi{\`e}re, Naman Goyal, Eric Hambro, Faisal Azhar, et~al. 2023{\natexlab{a}}.
\newblock Llama: Open and efficient foundation language models.
\newblock \emph{arXiv preprint arXiv:2302.13971}.

\bibitem[{Touvron et~al.(2023{\natexlab{b}})Touvron, Martin, Stone, Albert, Almahairi, Babaei, Bashlykov, Batra, Bhargava, Bhosale et~al.}]{llama2_23}
Hugo Touvron, Louis Martin, Kevin Stone, Peter Albert, Amjad Almahairi, Yasmine Babaei, Nikolay Bashlykov, Soumya Batra, Prajjwal Bhargava, Shruti Bhosale, et~al. 2023{\natexlab{b}}.
\newblock Llama 2: Open foundation and fine-tuned chat models.
\newblock \emph{arXiv preprint arXiv:2307.09288}.

\bibitem[{Wang et~al.(2022)Wang, Wei, Schuurmans, Le, Chi, Narang, Chowdhery, and Zhou}]{wang2022self}
Xuezhi Wang, Jason Wei, Dale Schuurmans, Quoc Le, Ed~Chi, Sharan Narang, Aakanksha Chowdhery, and Denny Zhou. 2022.
\newblock Self-consistency improves chain of thought reasoning in language models.
\newblock \emph{arXiv preprint arXiv:2203.11171}.

\bibitem[{Wang et~al.(2023{\natexlab{a}})Wang, Wei, Schuurmans, Le, Chi, Narang, Chowdhery, and Zhou}]{wang2023selfconsistency}
Xuezhi Wang, Jason Wei, Dale Schuurmans, Quoc~V Le, Ed~H. Chi, Sharan Narang, Aakanksha Chowdhery, and Denny Zhou. 2023{\natexlab{a}}.
\newblock Self-consistency improves chain of thought reasoning in language models.
\newblock In \emph{ICLR}.

\bibitem[{Wang et~al.(2023{\natexlab{b}})Wang, Zhang, Wang, Bhattacharya, Fu, and Wu}]{wang2023vaquita}
Yizhou Wang, Ruiyi Zhang, Haoliang Wang, Uttaran Bhattacharya, Yun Fu, and Gang Wu. 2023{\natexlab{b}}.
\newblock Vaquita: Enhancing alignment in llm-assisted video understanding.
\newblock \emph{arXiv preprint arXiv:2312.02310}.

\bibitem[{Wei et~al.(2022)Wei, Wang, Schuurmans, Bosma, Xia, Chi, Le, Zhou et~al.}]{wei2022chain}
Jason Wei, Xuezhi Wang, Dale Schuurmans, Maarten Bosma, Fei Xia, Ed~Chi, Quoc~V Le, Denny Zhou, et~al. 2022.
\newblock Chain-of-thought prompting elicits reasoning in large language models.
\newblock \emph{Advances in Neural Information Processing Systems}, 35:24824--24837.

\bibitem[{xjdr alt(2024)}]{entropix2024}
xjdr alt. 2024.
\newblock Entropix: Entropy based sampling and parallel cot decoding.
\newblock \url{https://github.com/xjdr-alt/entropix}.
\newblock Accessed: 2025-07-21.

\bibitem[{Xu et~al.(2024)Xu, Jin, Hao, Song, Sun, and Yuan}]{xu2024llava}
Guowei Xu, Peng Jin, Li~Hao, Yibing Song, Lichao Sun, and Li~Yuan. 2024.
\newblock Llava-o1: Let vision language models reason step-by-step.
\newblock \emph{arXiv preprint arXiv:2411.10440}.

\bibitem[{Yang et~al.(2024)Yang, Yang, Zhang, Hui, Zheng, Yu, Li, Liu, Huang, Wei et~al.}]{yang2024qwen2}
An~Yang, Baosong Yang, Beichen Zhang, Binyuan Hui, Bo~Zheng, Bowen Yu, Chengyuan Li, Dayiheng Liu, Fei Huang, Haoran Wei, et~al. 2024.
\newblock Qwen2. 5 technical report.
\newblock \emph{arXiv preprint arXiv:2412.15115}.

\bibitem[{Yao et~al.(2024)Yao, Yu, Zhao, Shafran, Griffiths, Cao, and Narasimhan}]{yao2024tree}
Shunyu Yao, Dian Yu, Jeffrey Zhao, Izhak Shafran, Tom Griffiths, Yuan Cao, and Karthik Narasimhan. 2024.
\newblock Tree of thoughts: Deliberate problem solving with large language models.
\newblock \emph{Advances in Neural Information Processing Systems}, 36.

\bibitem[{Yu et~al.(2023)Yu, Yang, Li, Wang, Lin, Liu, Wang, and Wang}]{yu2023mm}
Weihao Yu, Zhengyuan Yang, Linjie Li, Jianfeng Wang, Kevin Lin, Zicheng Liu, Xinchao Wang, and Lijuan Wang. 2023.
\newblock Mm-vet: Evaluating large multimodal models for integrated capabilities.
\newblock \emph{arXiv preprint arXiv:2308.02490}.

\bibitem[{Zhang et~al.(2025)Zhang, Bai, Wang, Wang, Dong, and Fu}]{zhang2025boosting}
Mingyuan Zhang, Yue Bai, Huan Wang, Yizhou Wang, Qihua Dong, and Yun Fu. 2025.
\newblock Boosting large language models with mask fine-tuning.
\newblock \emph{arXiv preprint arXiv:2503.22764}.

\bibitem[{Zhu et~al.(2024)Zhu, Chen, Shen, Li, and Elhoseiny}]{zhu2024minigpt}
Deyao Zhu, Jun Chen, Xiaoqian Shen, Xiang Li, and Mohamed Elhoseiny. 2024.
\newblock Mini{GPT}-4: Enhancing vision-language understanding with advanced large language models.
\newblock In \emph{ICLR}.

\end{thebibliography}

\clearpage
\appendix

\begin{center}
    \Large
    \textbf{Cautious Next Token Prediction}\\
    \vspace{0.5em}Supplementary Material \\
    \vspace{1.0em}
\end{center}

\section{Relationship of \ours with Beam Search~\citep{graves2012sequence} and Self-Consistency~\citep{wang2023selfconsistency}}
\label{apd: relation}
Traditional beam search maintains a fixed number of candidate expansions at every time step~\citep{graves2012sequence}. While effective in deterministic scenarios (e.g., speech recognition), beam search does not inherently adapt to model uncertainty. It can also over-penalize diverse continuations if beams converge to similar partial hypotheses.

By contrast, self-consistency~\citep{wang2023selfconsistency} generates multiple \emph{reasoning paths} and selects the most frequent final answer. Though powerful for reasoning tasks, it requires multiple full-path samples (often 5--40) regardless of how confident the model might be at intermediate steps. In essence, \emph{both} beam search and self-consistency can be seen as uniform multi-sample strategies.

\ours differs by linking sampling depth to real-time confidence signals via entropy, making it more cost-efficient. In cases of low uncertainty, \ours defaults to a simpler single-sample path akin to greedy decoding. As soon as the model’s internal distribution “spreads out,” \ours triggers more extensive branching and perplexity-based selection, thus combining the benefits of beam search’s path exploration with self-consistency’s final voting---but only where needed.

\section{Proof of Theorem~\ref{thm:heavy}}
\label{apd: proof}
\paragraph{Proof.}

\emph{(1) Full-sequence correctness.}  
Denote by $A_t$ the event that \emph{the token chosen at step $t$ is correct} and no previous errors occurred (so the partial sequence remains correct). For single-sample decoding, 
\begin{equation}
   P_{\text{Single}}(A_t) \;=\; 
      P_{\text{Single}}\bigl(w_t = c_t \mid A_1,\dots,A_{t-1}\bigr).
\end{equation}
If $H_t < H_{\min}$, then \ours also uses $N_t=1$ trial, so $P_{\text{CNTP}}(A_t) = P_{\text{Single}}(A_t)$.  
If $H_t \ge H_{\min}$, then by \ref{ass:confidence}, $p_{\theta}(c_t \mid s_{<t})$ is small, so a single sample might miss $c_t$. However, \ours uses $N_t>1$ trials. The probability that \emph{none} of these trials produce $c_t$ is $(1 - p_{\theta}(c_t\mid s_{<t}))^{N_t} \ll (1 - p_{\theta}(c_t\mid s_{<t}))$. Once $c_t$ is among the candidates, Assumption~\ref{ass:mono} ensures it will be selected due to the lowest perplexity. Therefore, for high-entropy steps,
\begin{align}
P_{\text{CNTP}}(A_t) &= 1 - \bigl(1 - p_{\theta}(c_t\mid s_{<t})\bigr)^{N_t}, \\
                    &> p_{\theta}(c_t\mid s_{<t}) = P_{\text{Single}}(A_t).
\end{align}

Hence, $P_{\text{CNTP}}(A_t)\ge P_{\text{Single}}(A_t)$ at every step, and strictly greater if $N_t>1$. The probability of the \emph{entire sequence} being correct is the product $\prod_{t=1}^L P(A_t)$. Thus $P_{\text{CNTP}}(\text{correct}) \ge P_{\text{Single}}(\text{correct})$ with strict inequality if any step used $N_t>1$.

\emph{(2) Expected cost.}  
Each decoding step entails $N_t$ forward passes if \ours chooses to sample $N_t$ times. Let $I_t = 1$ if $H_t \ge H_{\min}$ (high-entropy) and 0 otherwise. Then
\begin{equation}
\mathcal{C}_{\text{CNTP}}(S) = \sum_{t=1}^L N_t,
\end{equation}
where $N_t = 1 + (N_{\text{max}}-1)\,\mathbf{1}\bigl(I_t=1\bigr)
\quad\text{if } H_t \le H_{\max}$ or saturates at $N_{\text{max}}$ if $H_t>H_{\max}$ (we clamp $N_t\le N_{\text{max}}$). Let $p = \frac{1}{L}\sum_{t=1}^L \mathbb{E}[I_t]$ be the expected fraction of high-entropy steps. Then
\begin{align}
   \mathbb{E}\bigl[\mathcal{C}_{\text{CNTP}}(S)\bigr]
   &=
   \sum_{t=1}^L \mathbb{E}[N_t] \\
   &\le
   L\,\Bigl[ 1 + p\,(N_{\text{max}}-1)\Bigr].
\end{align}

This is strictly less than $LN_{\text{max}}$, which would be the cost of always sampling $N_{\text{max}}$ continuations at every step (uniform multi-sample decoding). 
$\qedsymbol$

\section{More Experiment Setting \& Details}
\label{apd: setting}

We tune the temperature value for the baseline methods and for \ours in list $[0.6, 0.8, 1.0, 1.2]$ and select the best performance. We tune the beam size in list $[2,3,5,10,20,40]$ and select the best performance for all experiments. For all the single-reasoning chain stochastic approaches in the LLM experiments, we run $5$ times independently and report the average performance and sample standard deviations. For the TruthfulQA experiments, we use Llama-2-7B as LLM judge to output the truth accuracy, info accuracy and truth info accuracy. For the GSM8K experiments, we adopt 9-shot CoT during inference. For StrategyQA dataset,we adopt 6-shot CoT during inference. For MATH dataset, we adopt 4-shot CoT during inference. All experiments are done on NVIDIA A100-SXM4-80GB GPUs and Intel(R) Xeon(R) Platinum 8275CL CPU @ 3.00GHz CPUs with $96$ logical processors.

\end{document}